\documentclass[10pt,twocolumn,letterpaper]{article}

\usepackage[pagenumbers]{iccv}   %

\PassOptionsToPackage{dvipsnames,table}{xcolor}

\usepackage{wrapfig}
\usepackage{colortbl}
\usepackage[cjk]{kotex}

\usepackage{lipsum}
\usepackage[normalem]{ulem}

\usepackage{graphicx}
\usepackage{multirow}
\usepackage{booktabs}
\usepackage{array}
\usepackage[ruled,linesnumbered]{algorithm2e}
\usepackage{amsmath,amssymb}
\usepackage{enumitem}

\usepackage{multirow}
\usepackage{makecell}

\definecolor{iccvblue}{rgb}{0.21,0.49,0.74}
\usepackage[pagebackref,breaklinks,colorlinks,allcolors=iccvblue]{hyperref}

\usepackage{xcolor}
\usepackage{listings}
\usepackage{algpseudocode}
\lstdefinestyle{pythonstyle}{
    language=Python,
    basicstyle=\ttfamily\footnotesize,
    keywordstyle=\color{keywordblue}\bfseries,
    commentstyle=\color{commentgreen},
    stringstyle=\color{stringred},
    breaklines=true,
    showstringspaces=false,
    tabsize=2
}
\definecolor{commentgreen}{rgb}{0,0.5,0}  %
\definecolor{keywordblue}{rgb}{0,0,0.8}   %
\definecolor{stringred}{rgb}{0.6,0,0}     %

\def\paperID{15} %
\def\confName{ICCV}
\def\confYear{2025}

\title{Can Synthetic Images Conquer Forgetting? \\ Beyond Unexplored Doubts in Few-Shot Class-Incremental Learning}

\author{%
  Junsu Kim$^{1,3}$ \quad Yunhoe Ku$^{2,3}$ \quad Seungryul Baek$^{1}$ \\\\
  $^{1}$UNIST \qquad
  $^{2}$DeepBrain AI \qquad
  $^{3}$\shortstack{NVIDIA Foundation Models LAB, MODULABS}%
}

\begin{document}
\maketitle{
}

\maketitle
\begin{abstract}
Few-shot class-incremental learning (FSCIL) remains challenging due to severe data scarcity and persistent threat of catastrophic forgetting. While generative replay with synthetic images has emerged as a promising approach, critical questions regarding synthetic data usage—such as the optimal quantity per class, generation strategy, and timing of integration—remain largely unexplored. In this work, we conduct an extensive, rigorously controlled analysis to systematically uncover subtle yet impactful factors influencing synthetic replay effectiveness in FSCIL. Through comprehensive experiments on CUB-200 and \textit{mini}ImageNet, we demonstrate that optimization-based methods, particularly the Textual Inversion strategy, excel by embedding detailed class-specific semantic information. Furthermore, we reveal dataset-dependent trade-offs related to the timing of synthetic image integration during base training. Our findings provide clear, trustworthy insights into previously overlooked aspects of synthetic images, establishing a foundational step for future advancements in FSCIL.
\end{abstract}

\section{Introduction}
\label{sec:introduction}

\begin{flushleft}
\textit{“The only limit to our realization of tomorrow will be our doubts of today.”}
\hfill\small—\,Franklin~D.\ Roosevelt
\end{flushleft}

\vspace{0.1em}
In a rapidly evolving field of artificial intelligence, continual learning (CL) has emerged as a crucial paradigm to enable models to incrementally acquire and integrate new knowledge without retraining from scratch. However, our uncertainties—or doubts—about effectively managing past and new knowledge integration have persistently hindered realizing the full potential of CL. Among these doubts, the phenomenon known as catastrophic forgetting remains a challenge, wherein newly acquired information disruptively overwrites previously learned knowledge. To this end, traditional strategies, such as exemplar replay~\cite{rebuffi2017iCaRL, kim2024sddgr, kim2024classwise, shi2021overcoming_replay3}, regularization~\cite{akyurek2022Regularizer, lopez2017gradient-reg, kirkpatrick2017overcoming-reg}, and knowledge distillation~\cite{rebuffi2017iCaRL, kim2024sddgr, zhao2023BiDistFSCIL, simon2021learning-dist, kim2024vlmpl}, have partially alleviated these issues across various tasks (\eg, detection, classification, segmentation). However, these methods typically assume abundant data availability at each incremental session—an idealistic condition rarely encountered in real-world scenarios.

Recognizing such practical limitations, few-shot class-incremental learning (FSCIL) has recently gained significant attention. FSCIL introduces classes incrementally with extremely limited examples per class, forcing inherent challenges like catastrophic forgetting or overfitting to a few samples. Consequently, the models often struggle significantly to consolidate new knowledge without compromising previously learned information. Despite significant efforts~\cite{song2023SAVC, ahmed2024orco, kim2025diffusionfscil, yang2023NC_FSCIL, tang2024yourself} to address these complexities, fully resolving them remains an open research direction that continues to challenge the community.
\begin{figure}[t!]
    \centering
    \includegraphics[width=1.0\linewidth]{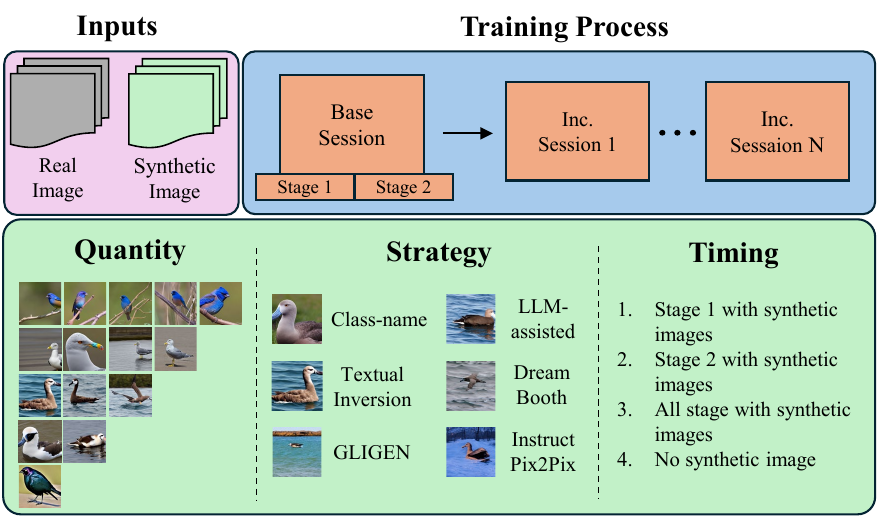}
    \vspace{-1em}
    \caption{\textbf{Overview of our analysis framework.} We systematically investigate three critical yet previously overlooked factors influencing the effectiveness of synthetic image replay in FSCIL: (1) the number of synthetic images per class (\textit{Quantity}), (2) various synthetic image generation methods (\textit{Strategy}), and (3) the timing of synthetic image integration within base-session training (\textit{Timing}). Specifically, the \textit{Timing} analysis examines how different integration schedules for synthetic images during base-session training affect the backbone’s representational diversity, consequently impacting incremental learning performance.}
    \label{fig:overview}
    \vspace{-1em}
\end{figure}

Among the numerous strategies proposed to overcome FSCIL challenges, replay methods have proven particularly effective so that they are widely adopted in diverse CL scenarios. Such approaches mitigate catastrophic forgetting by periodically revisiting exemplars from past sessions. However, replay methods inherently face strict memory constraints, redundancy, and diminishing returns over incremental steps. Crucially, practical concerns such as a privacy issue~\cite{meng2024diffclass, shin2017DGR} have motivated recent research toward generative replay strategies~\cite{kim2024sddgr, jodelet2023DiffusionImageReplayCIL, meng2024diffclass, gao2023ddgr, liu2022DataFreeReplay, agarwal2022semantics, meng2024diffclass, shin2017DGR}. Among these, diffusion-based generative models, notably Stable Diffusion (SD)~\cite{Rombach_2022LDM}, have received considerable attention due to their superior capabilities in synthesizing diverse, high-quality images—essential for effective replay.

Despite their promising results, existing diffusion-based replay methods often lack in-depth insights into the precise impact of synthetic images on the context of CL performance. Consequently, a fundamental and intriguing question naturally arises: \textit{How exactly do synthetic images influence FSCIL performance?} To address this question comprehensively, our study rigorously investigates synthetic image strategies by meticulously controlling variables and examining three critical aspects, as illustrated in \cref{fig:overview}: 1) the quantity of synthetic images per class (\textit{Quantity}), 2) a comparative evaluation of various generative strategies (\textit{Strategy}), and 3) the timing of synthetic image integration within the base session training (\textit{Timing}). Notably, the timing analysis explores previously unexamined scenarios where synthetic images are employed not only as replay exemplars but also strategically combined with real images during base-session training to enhance backbone representational diversity, similar in spirit to augmentation approaches used in prior FSCIL methods~\cite{ahmed2024orco, kim2025diffusionfscil, song2023SAVC, peng2022ALICE, zhou2022FACT} (see \cref{tab:experiment_settings} for details).

Through systematic and fair evaluations, we aim to clarify the significance of synthetic images, highlighting previously unexplored controllable factors that critically influence their effectiveness in FSCIL. Ultimately, our analysis not only resolves existing uncertainties but also provides clear evidence supporting broader and more confident adoption of synthetic images within FSCIL—\textit{serving as a foundational step enabling future research to progress confidently toward broader horizons, unburdened by previous doubts}.
\begin{figure*}[t!]
    \centering
    \includegraphics[width=1.00\linewidth]{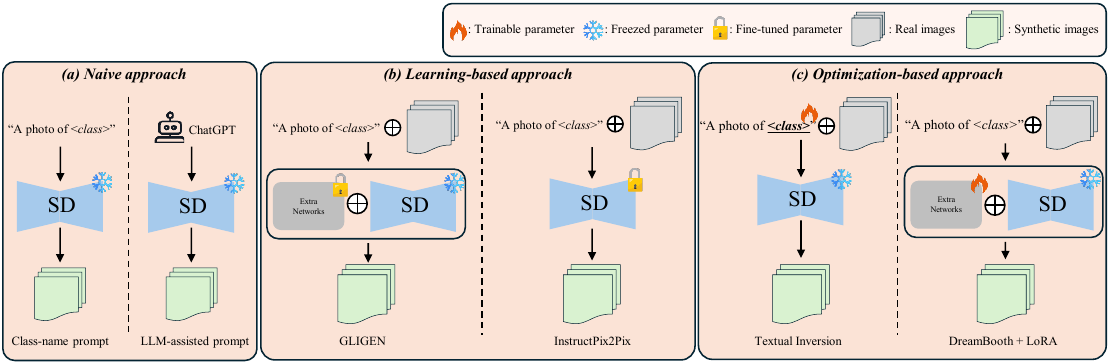}
    \vspace{-1em}
    \caption{\textbf{Illustration of the generative strategies utilized in this study.} We categorize generative methods into three distinct approaches: \textbf{(a) Na\"ive approach}, directly utilizing a frozen SD model with simple text prompts (\eg, Class-name or LLM-assisted prompts); \textbf{(b) Learning-based approach}, employing external pretrained networks with additional networks (\eg, GLIGEN and InstructPix2Pix) to guide or condition a frozen SD model; and \textbf{(c) Optimization-based approach}, where small portions of the SD model (\eg, text embeddings in Textual Inversion or LoRA parameters in DreamBooth) are explicitly optimized to generate class-specific synthetic images.}
    \label{fig:various_strategies}
    \vspace{-1em} 
\end{figure*}

\begin{table}[t]
    \centering
    \caption{\textbf{Variation of training strategies according to dataset type.} We define four setups based on different timing for integrating synthetic (\textbf{S}) and real (\textbf{R}) images during base-session training. $\otimes$ denotes that the second training stage is not applied.}
    \label{tab:experiment_settings}
    \resizebox{\linewidth}{!}{%
    \begin{tabular}{cc|c}
        \toprule
        \multicolumn{2}{c|}{\textbf{Base session training}} & \multirow{2}{*}{\textbf{Notation}} \\
        \textbf{Stage 1} & \textbf{Stage 2} & \\ \midrule
        \textbf{R}eal & $\otimes$ & \textbf{R} \textit{(full)} \\
        \textbf{R}eal + \textbf{S}ynthetic & $\otimes$ & \textbf{R + S} \textit{(full)} \\
        \textbf{R}eal & \textbf{R}eal + \textbf{S}ynthetic & \textbf{R} \textit{(1st)} $\rightarrow$ \textbf{R+S} \textit{(2nd)} \\
        \textbf{R}eal + \textbf{S}ynthetic & \textbf{R}eal & \textbf{R+S} \textit{(1st)} $\rightarrow$ \textbf{R} \textit{(2nd)} \\
        \bottomrule
    \end{tabular}}
\end{table}

\section{Related Work}
\label{sec:related_work}
\subsection{Diffusion models and their applications}
Large-scale text-to-image (T2I) diffusion models, such as DALL-E 2~\cite{ramesh2021Dalle}, Imagen~\cite{saharia2022imagen}, and Stable Diffusion (SD)~\cite{Rombach_2022LDM}, have demonstrated remarkable capabilities in generating photorealistic and semantically coherent images from natural-language prompts. Recent research predominantly focuses on adapting these pretrained models rather than training them from scratch, aiming to enhance image fidelity, controllability, and alignment with user intent. For instance, DreamBooth~\cite{ruiz2023dreambooth} fine-tunes the entire diffusion model using a small set of images to personalize generation towards specific visual concepts, while Textual Inversion~\cite{gal2022textual_inversion} optimizes textual embeddings to capture specific image-level concepts within the pretrained embedding space. Additionally, GLIGEN~\cite{li2023gligen} introduces extra parameters into SD, enabling spatial control via grounding boxes, and InstructPix2Pix~\cite{brooks2023instructpix2pix} fine-tunes the diffusion model for guided image editing from textual instructions. However, despite extensive exploration of adapted diffusion models in broader CL settings, the explicit impact of using synthetic images exclusively—without additional mechanisms—remains largely unexplored, leaving critical uncertainties unaddressed.

\subsection{Few-shot class-incremental learning}
\label{subsec:rw_fscil}
Few-shot class-incremental learning (FSCIL)~\cite{zhang2023fewshotsurvey,tian2024survey} extends the class-incremental learning (CIL) paradigm~\cite{cauwenberghs2000incremental,li2017learning,rebuffi2017iCaRL} to scenarios where new classes arrive with only a handful of examples, often causing severe catastrophic forgetting and overfitting. Common approaches~\cite{zhou2022FACT,peng2022ALICE,zhao2021mgsvf,yang2023NC_FSCIL,tang2024yourself,ahmed2024orco,song2023SAVC} involve developing robust prototypes or feature representations that balance previous and novel knowledge. Some methods employ meta-learning~\cite{zhu2021self_promoted,chi2022MetaFSCIL,chen2021IDLVQ,zou2022CLOM,hersche2022C_FSCIL} to quickly adapt to new classes, while others leverage replay strategies~\cite{liu2022DataFreeReplay,agarwal2022semantics,cheraghian2021semantic,dong2021few, rebuffi2017iCaRL}, utilizing small exemplar sets or synthetic data generated by additional generative models. Given the severely limited availability of real data, effectively utilizing synthetic images emerges as a pivotal strategy to alleviate catastrophic forgetting and overfitting, core challenges identified in FSCIL.

\subsection{Diffusion-based methods for CL}
Building upon the powerful capabilities of diffusion models, recent incremental learning studies have actively incorporated T2I methods to generate synthetic replay exemplars. Diffusion models have demonstrated promising results across various CL scenarios, including class-specific replay~\cite{meng2024diffclass,jodelet2023DiffusionImageReplayCIL,kim2025diffusionfscil}, object detection replay~\cite{kim2024sddgr}, and future-class pretraining~\cite{jodelet2025futureproofing}. Similarly, GAN-based strategies have been frequently adopted to generate synthetic exemplars for FSCIL tasks~\cite{agarwal2022semantics,liu2022DataFreeReplay}. Nevertheless, existing analyses remain superficial, overlooking critical factors such as exemplar quantity, generative strategy selection, and integration timing. To address these gaps, we systematically evaluate diverse diffusion-based generative strategies under controlled conditions, offering detailed insights to guide their confident and informed future adoption in FSCIL.

\section{Preliminary}
\label{sec:preliminaries}
\subsection{Few-shot class incremental learning}
Few-shot class incremental learning (FSCIL) aims to continuously train neural networks to recognize novel classes introduced with an extremely limited number of images, while simultaneously maintaining knowledge on previously learned classes. A conventional FSCIL protocol involves a series of training sessions, represented as $\{\mathcal{S}_0, \mathcal{S}_1, \dots, \mathcal{S}_T \}$. In the base session $\mathcal{S}_0$, the model is trained on a relatively large dataset $\mathcal{D}_0$, containing multiple base classes $\mathcal{C}_0$. During incremental sessions $\mathcal{S}_t$ (for $t \geq 1$), new classes $\mathcal{C}_t$ are introduced through significantly smaller, few-shot datasets $\mathcal{D}_t$.

To mitigate catastrophic forgetting, existing methods~\cite{kim2025diffusionfscil, ahmed2024orco, yang2023NC_FSCIL, zhao2023BiDistFSCIL} commonly maintain an exemplar buffer $\mathcal{B}_\text{real}$ containing selected real samples from earlier sessions. Some approaches~\cite{agarwal2022semantics, liu2022DataFreeReplay} further enhance this strategy by incorporating synthetic exemplars $\mathcal{B}_\text{gen}$ generated using generative models. Consequently, the training set at incremental sessions $\mathcal{S}_t$ combines the newly introduced few-shot samples with exemplars from both real and synthetic sources, formally represented as: $\mathcal{D}_t' = \mathcal{D}_t \cup \mathcal{B}_\text{real} \cup \mathcal{B}_\text{gen}$. 

Performance at each incremental session is cumulatively evaluated over all classes introduced up to the current session, formally defined as: $\mathcal{C}_{all} = \bigcup_{i=0}^{t} \mathcal{C}_i$.

\subsection{Text-to-image diffusion models}
Stable Diffusion (SD)~\cite{Rombach_2022LDM} is a prominent text-conditioned generative model capable of synthesizing high-quality images from a text prompt $\mathbf{p}$. During the image-generation process, SD starts from pure Gaussian noise $\mathbf{z}_\text{T}$ and progressively removes noise guided by the text prompt, ultimately producing latent representations $\mathbf{z}_0$ reflecting the intended semantics. The denoising procedure employs a UNet~\cite{ronneberger2015unet} integrated with cross-attention~\cite{vaswani2017attention}, enabling conditional image generation via a textual embedding $\mathbf{w}$ obtained by encoding the input text prompt using a pretrained text encoder~\cite{radford2021CLIP}. The latent representation $\mathbf{z}_0$ is then decoded into the image domain through a variational autoencoder (VAE) decoder~\cite{esser2021VQVAE, kingma2013VAE}, typically yielding synthetic images with a resolution of $512\times512$ pixels.

\section{Method} 
\label{section:method}

Our goal is to systematically analyze the impact of synthetic images, generated through diverse strategies, in alleviating catastrophic forgetting within FSCIL scenarios. To facilitate clear understanding and structured evaluation, we categorize these generation strategies into three distinct groups: (1) \emph{Na\"ive approaches}, relying solely on straightforward prompting methods without additional training or modifications; (2) \emph{Learning-based approaches}, employing auxiliary pretrained networks to guide image generation; and (3) \emph{Optimization-based approaches}, directly optimizing or fine-tuning generative models to capture detailed, class-specific visual concepts.

\cref{fig:various_strategies} visually summarizes these synthetic generation strategies. Next, we detail each strategy and its integration into FSCIL training.
\subsection{Na\"ive approach}
\noindent\textbf{Class-name prompt.}
The class-name prompt method directly leverages SD without additional fine-tuning or modifications by employing straightforward text prompt engineering. Following the template of SDDR~\cite{jodelet2023DiffusionImageReplayCIL}, we simply formulate prompts as \small\texttt{"A photo of \{class-name\}"} under the assumption that the exact class-name of previously learned class is known. This approach provides a simple yet effective baseline for generating synthetic images aligned with their respective class semantics.

\noindent\textbf{LLM-assisted prompt.}
To further enrich the class-name prompt without additional training costs, we consider utilizing world-scale large language models (LLMs), such as ChatGPT-o3~\cite{openai_chatgpt_o3}, to augment the prompt with additional details. Specifically, the LLM generates concise yet informative text descriptions tailored to enhance the basic class-name prompts, structured as \small\texttt{"A photo of \{class-name\}, \{description\}"}. For example, given the class-name ``goose," the LLM may produce a description such as ``a large waterbird with a long neck and webbed feet, typically found near lakes and rivers," which is then appended directly to the original prompt. Such additional context significantly improves the semantic richness of generated synthetic images, thereby enhancing their effectiveness in FSCIL replay scenarios.

\subsection{Learning-based approach}
\noindent\textbf{GLIGEN.}
We employ GLIGEN~\cite{li2023gligen}, a model enabling spatially user-intended image synthesis by integrating grounding inputs, such as bounding boxes and image embeddings, into pretrained SD models through gated self-attention layers. Inspired by SDDGR~\cite{kim2024sddgr}, which effectively utilizes GLIGEN-based replay strategy for CL of object detection, we utilize bounding box coordinates, image embeddings, and text prompts to additional grounding network for explicitly guiding SD generation. Specifically, we construct text prompt using class-name information (\eg, \small\texttt{"A photo of \{class-name\}"}), utilize ground-truth bounding box (\eg, [x, y, w, h]) indicating precise bird location, and employ image embeddings to align the synthesized image fidelity closely with the corresponding reference images.

\noindent\textbf{InstructPix2Pix.}
InstructPix2Pix~\cite{brooks2023instructpix2pix} is an image editing model that fine-tunes a pretrained SD model to follow editing instruction provided as text prompts. Given input images and corresponding editing text prompt (\eg, ``change the bird's color to blue"), the model generates an edited image aligned semantically and visually with the provided instruction. To leverage this capability in our setting, we first identify a set of effective editing instructions that modify only the target regions while leaving other areas unchanged. Then, each instruction is selected with a certain probability to construct the overall editing prompt. Finally, we then apply that overall instruction to available few-shot reference images, synthesizing diverse replay exemplars explicitly tailored for our work.

\subsection{Optimization-based approach}
\noindent\textbf{Textual Inversion.}
Inspired by the recent study~\cite{kim2025diffusionfscil}, we utilize the Textual Inversion~\cite{gal2022textual_inversion}, which optimizes class-specific textual embeddings. Specifically, for each class-name, we introduce a unique special prompt $\textbf{p}^*$ and individually optimize the corresponding textual embedding $\mathbf{w}^*$ derived from the CLIP~\cite{radford2021CLIP} text encoder, using only the available few-shot examples. These optimized embeddings $\mathbf{w}^*$, effectively capturing the distinctive visual concepts of their respective classes, are pre-computed and stored. During image generation, we retrieve these embeddings by incorporating the special prompts $\mathbf{p}^*$ into a structured template (\eg, \texttt{"A photo of \{$\mathbf{p}^*$\}"}), thus synthetic images that precisely embody each few-shot class concept.

\vspace{0.3em}
\noindent\textbf{DreamBooth with LoRA.}
Similar to Textual Inversion, we utilize a lightweight variant of DreamBooth~\cite{ruiz2023dreambooth} that fine‑tunes small Low-Rank Adaptation (LoRA) modules~\cite{hu2022lora} integrated into the frozen UNet of SD, instead of updating the entire SD. Specifically, for each previously learned class $\text{c} \in \mathcal{C}_{<t}$, we allocate a unique identifier token \texttt{[V\textsubscript{c}]}, whose embedding is appended to the class-name position within a fixed prompt template (\eg, \texttt{"A photo of [V\textsubscript{c}]}"). Then, utilizing the same few-shot images available per class as in Textual Inversion, we fine-tune the LoRA modules attached to the frozen UNet and store the resulting LoRA weights as compact, class-specific parameter files. During image generation, we simply load the stored LoRA parameters for the desired class and input the associated predefined prompt, thereby synthesizing images that faithfully embody the visual characteristics of each class.

\begin{table*}[t]
\centering
\begin{minipage}[t]{0.53\textwidth}
\centering
\captionof{table}{Session-wise FSCIL accuracy (\%) with different synthetic-image generation strategies on the CUB-200 dataset. AA denotes average accuracy over all sessions. All methods use 5 images per class; every other experimental setting is identical. \textbf{Bold} and \underline{underlined} values denote best and second-best per session.}
\label{tab:cub_acc}
\setlength{\tabcolsep}{2pt}
\renewcommand{\arraystretch}{1.0}

\resizebox{\linewidth}{!}{%
\begin{tabular}{l|cccccccccccc}
\toprule
\multirow{2}{*}{\textbf{Synthetic Method}} & \multicolumn{12}{c}{\textbf{CUB-200 Session Accuracy (\%)}} \\
\cmidrule{2-13}
& 0 & 1 & 2 & 3 & 4 & 5 & 6 & 7 & 8 & 9 & 10 & AA \\
\midrule
No synthetic image  & \textbf{80.37} & 59.21 & 67.78 & 62.58 & 60.85 & 56.56 & 53.99 & 51.03 & 48.14 & 48.65 & 49.64 & 58.07 \\
\midrule
\rowcolor{gray!20}\multicolumn{13}{l}{\textbf{Na\"ive approach}}\\
Class-name prompt  & \textbf{80.37} & 70.44 & 65.39 & 61.21 & 58.14 & 55.78 & 54.55 & 51.80 & 49.79 & 49.03 & 48.36 & 58.62 \\
LLM-assisted~\cite{openai_chatgpt_o3} & \textbf{80.37} & \underline{71.11} & 66.18 & 62.63 & 60.20 & 57.17 & 55.96 & 52.90 & 50.54 & 50.56 & 49.09 & 59.70 \\
\midrule
\rowcolor{gray!20}\multicolumn{13}{l}{\textbf{Learning-based approach}}\\
GLIGEN~\cite{li2023gligen}  & \textbf{80.37} & 71.24 & 66.21 & 62.50 & 58.84 & 56.29 & 53.90 & 51.68 & 48.69 & 48.80 & 47.67 & 58.74 \\
InstructPix2Pix~\cite{brooks2023instructpix2pix} & \textbf{80.37} & 71.91 & 67.06 & 63.57 & 60.23 & 56.66 & 55.61 & 52.92 & 50.27 & 50.04 & 48.26 & 59.72 \\
\midrule
\rowcolor{gray!20}\multicolumn{13}{l}{\textbf{Optimization approach}}\\
Textual Inversion~\cite{gal2022textual_inversion} & \textbf{80.37} & \textbf{72.99} & \textbf{69.07} & \textbf{66.31} & \textbf{63.60} & \textbf{61.56} & \textbf{59.41} & \underline{57.56} & \underline{55.69} & \textbf{55.75} & \underline{55.28} & \textbf{63.42} \\
DreamBooth~\cite{ruiz2023dreambooth} & \textbf{80.37} & \textbf{72.99} & \textbf{69.07} & \underline{64.78} & \underline{62.66} & \underline{61.12} & \underline{59.28} & \textbf{57.85} & \textbf{55.80} & \underline{55.26} & \textbf{55.37} & \underline{63.14} \\
\bottomrule
\end{tabular}}
\end{minipage}
\hspace{0.0005\textwidth}
\begin{minipage}[t]{0.46\textwidth}
\centering
\captionof{table}{Session-wise FSCIL accuracy (\%) with different synthetic-image generation strategies on \textit{mini}ImageNet. AA = average accuracy over all sessions. All methods use 5 images per class; every other experimental setting is identical. \textbf{Bold} and \underline{underlined} values denote best and second-best per session.}
\label{tab:mini_acc}
\setlength{\tabcolsep}{2pt}
\renewcommand{\arraystretch}{1.0}

\resizebox{\linewidth}{!}{%
\begin{tabular}{l|cccccccccc}
\toprule
\multirow{2}{*}{\textbf{Synthetic Method}} &
\multicolumn{10}{c}{\textbf{\textit{mini}ImageNet Session Accuracy (\%)}} \\
\cmidrule(lr){2-11}
& 0 & 1 & 2 & 3 & 4 & 5 & 6 & 7 & 8 & AA \\
\midrule
No synthetic image & 68.70 & 61.11 & 54.86 & 52.21 & \underline{49.71} & 44.94 & 42.69 & 41.46 & 38.40 & 50.45 \\
\midrule
\rowcolor{gray!20}\multicolumn{11}{l}{\textbf{Naïve approach}}\\
Class-name & 68.70 & \underline{62.09} & \textbf{57.07} & 52.60 & 49.29 & 45.91 & 43.53 & 41.63 & \underline{39.92} & 51.19 \\
LLM-assisted~\cite{openai_chatgpt_o3} & 68.70 & \textbf{62.35} & 55.61 & 51.81 & 48.29 & 45.08 & 43.03 & 41.08 & 39.64 & 50.62 \\
\midrule
\rowcolor{gray!20}\multicolumn{11}{l}{\textbf{Learning-based approach}}\\
GLIGEN~\cite{li2023gligen} & 68.70 & 61.98 & 56.50 & 52.00 & 48.92 & 45.65 & 43.17 & 41.36 & 38.91 & 50.80 \\
InstructPix2Pix~\cite{brooks2023instructpix2pix} & 68.70 & 61.46 & 55.51 & 51.71 & 48.56 & 45.14 & 43.09 & 41.31 & 39.19 & 50.52 \\
\midrule
\rowcolor{gray!20}\multicolumn{11}{l}{\textbf{Optimization approach}}\\
Textual Inversion~\cite{gal2022textual_inversion} & 68.70 & 61.89 & \underline{56.90} & \textbf{53.24} & \textbf{50.08} & \textbf{47.05} & \textbf{44.84} & \textbf{42.58} & \textbf{40.50} & \textbf{51.75} \\
DreamBooth~\cite{ruiz2023dreambooth} & 68.70 & 61.92 & 56.41 & \underline{52.71} & 49.25 & \underline{46.19} & \underline{44.06} & \underline{41.94} & 39.72 & \underline{51.21} \\
\bottomrule
\end{tabular}}
\end{minipage}

\end{table*}

\section{Experiment}
\label{sec:experiment}

\subsection{Setup}
\label{subsec:setup}
\noindent\textbf{Datasets.}
We conduct experiments on two widely-used benchmarks: the fine-grained dataset CUB-200~\cite{wah2011CUB200}, comprising 11,788 images of 200 bird species at a resolution of $224\times224$ pixels; and \emph{mini}ImageNet~\cite{russakovsky2015imagenetmini}, containing 100 classes, each with 500 training and 100 test images at a resolution of $84\times84$ pixels. Following the standard FSCIL evaluation protocol~\cite{yang2023NC_FSCIL, song2023SAVC, peng2022ALICE, kim2025diffusionfscil}, we use 100 base classes followed by 10 incremental sessions (\ie, 10-way 5-shot) for CUB-200, and 60 base classes followed by 8 incremental sessions (\ie, 5-way 5-shot) for \emph{mini}ImageNet.

\noindent\textbf{Evaluation metrics.}
We utilize two primary metrics: (i) \emph{session accuracy}, measuring classification performance at each incremental session, and (ii) \emph{average accuracy} (AA), computed as the mean accuracy across all sessions up to the current session, thus reflecting the model's consistent recognition of both previously learned and newly introduced classes. Additionally, for deeper analysis, we measure \emph{performance drop} (PD), defined as the accuracy difference between the initial (Base) and final (Last) sessions.

\subsection{Implementation details.}
To rigorously assess the impact of synthetic images, we explicitly outline the common implementation details and specific setups used for each generative strategy.

\noindent\textbf{Common setup.}
We employ pre-trained SD v1.5~\cite{Rombach_2022LDM} unless otherwise specified, using a classifier-free guidance scale of 7.5~\cite{ho2022classifier_free_guidance} and the DDIM scheduler~\cite{song2020ddim}, without negative prompts across all experiments. For the CUB-200, following prior approaches~\cite{yang2023NC_FSCIL, tang2024yourself, chi2022MetaFSCIL, peng2022ALICE, zhao2023BiDistFSCIL}, we utilize ImageNet~\cite{deng2009imagenet} as a pre-training dataset. In line with established replay methods~\cite{rebuffi2017iCaRL, zhao2023BiDistFSCIL, dong2021few, shi2021overcoming_replay3, ahmed2024orco}, we maintain \textbf{one real exemplar per class} in the buffer $\mathcal{B}_\text{real}$. Throughout our experiments, we utilize a ResNet-18~\cite{he2016resnet} backbone with synchronized batch normalization and a two-layer MLP connected to the classifier. We implement a pre-defined prototype~\cite{yang2023NC_FSCIL} as our classifier. The backbone network is trained only during the base session ($\mathcal{S}_0$) and remains frozen thereafter ($S_{t \geq 1}$).

For training, we use the AdamW optimizer~\cite{loshchilov2017AdamW} with dataset-specific learning rates: 1e-3 for CUB-200 and 3e-3 for \emph{mini}ImageNet, alongside a weight decay of 5e-4 and a cosine learning rate scheduler. Our data augmentation pipeline includes horizontal flipping, color jittering, and random erasing. Regarding batch size, we utilize 320 for CUB-200 and 640 for \emph{mini}ImageNet during the base session, while uniformly setting it to 80 in all incremental sessions for both datasets. In incremental sessions, training begins with 105 iterations for session $\mathcal{S}_1$, incrementally adding 10 iterations for each subsequent session. All experiments are conducted using 5 RTX 3090 GPUs.

\noindent\textbf{Textual Inversion.}
Following~\citet{kim2025diffusionfscil}, we set the length of the optimized textual embedding vectors to 5 for CUB-200 and 7 for \emph{mini}ImageNet, as these values were empirically determined to yield optimal FSCIL performance. We optimize these embeddings using only the available few-shot examples for each specific class.

\noindent\textbf{DreamBooth with LoRA.}
We implement DreamBooth~\cite{ruiz2023dreambooth} with LoRA~\cite{hu2022lora} using the diffusers~\cite{von-platen-etal-2022-diffusers} library, adopting the library's recommended hyperparameters and keeping all remaining parameters identical to the original settings. Specifically, we fine-tune only the LoRA modules attached to the frozen UNet, leaving all other model components fixed. During optimization, we combine a class-specific prior-preservation loss (with a weight of 1.0) and a reconstruction loss (MSE), consistent with the original DreamBooth configuration.

\noindent\textbf{GLIGEN.} 
We use ground-truth bounding boxes ([x, y, w, h]) from CUB-200, while extracting bounding boxes for \emph{mini}ImageNet using YOLOv8~\cite{Jocher_YOLO_by_Ultralytics_2023}. Additionally, we obtain image embeddings via the CLIP image encoder for both datasets to guide the image synthesis process. Unfortunately, since the authors of GLIGEN have not publicly released pretrained weights based on SD v1.5, we utilize their available pretrained weights built upon SD v1.4.

\noindent\textbf{InstructPix2Pix.}
We design four prompt templates to introduce variations without altering the identity of the object: \small\texttt{"change season to \{season\}"}, \small\texttt{"add \{item\} to the \{class\}"}, \small\texttt{"the \{class\} moves \{speed\}ly"}, and \small\texttt{"make it \{time\}"}. For each placeholder, we select from predefined options: \{season\} from [spring, summer, fall, winter], \{item\} from [hat, scarf, glasses, necklace], \{speed\} from [slow, fast], and \{time\} from [morning, afternoon, night]. Please note that, for flexible image generation, we set classifier-free guidance of input image as 1.

\section{Analysis}
\label{section:analysis}

\subsection{Effect of the number of synthetic images}
\label{subsection:number_of_images}
\begin{figure}[t]
    \centering
    \includegraphics[width=\linewidth]{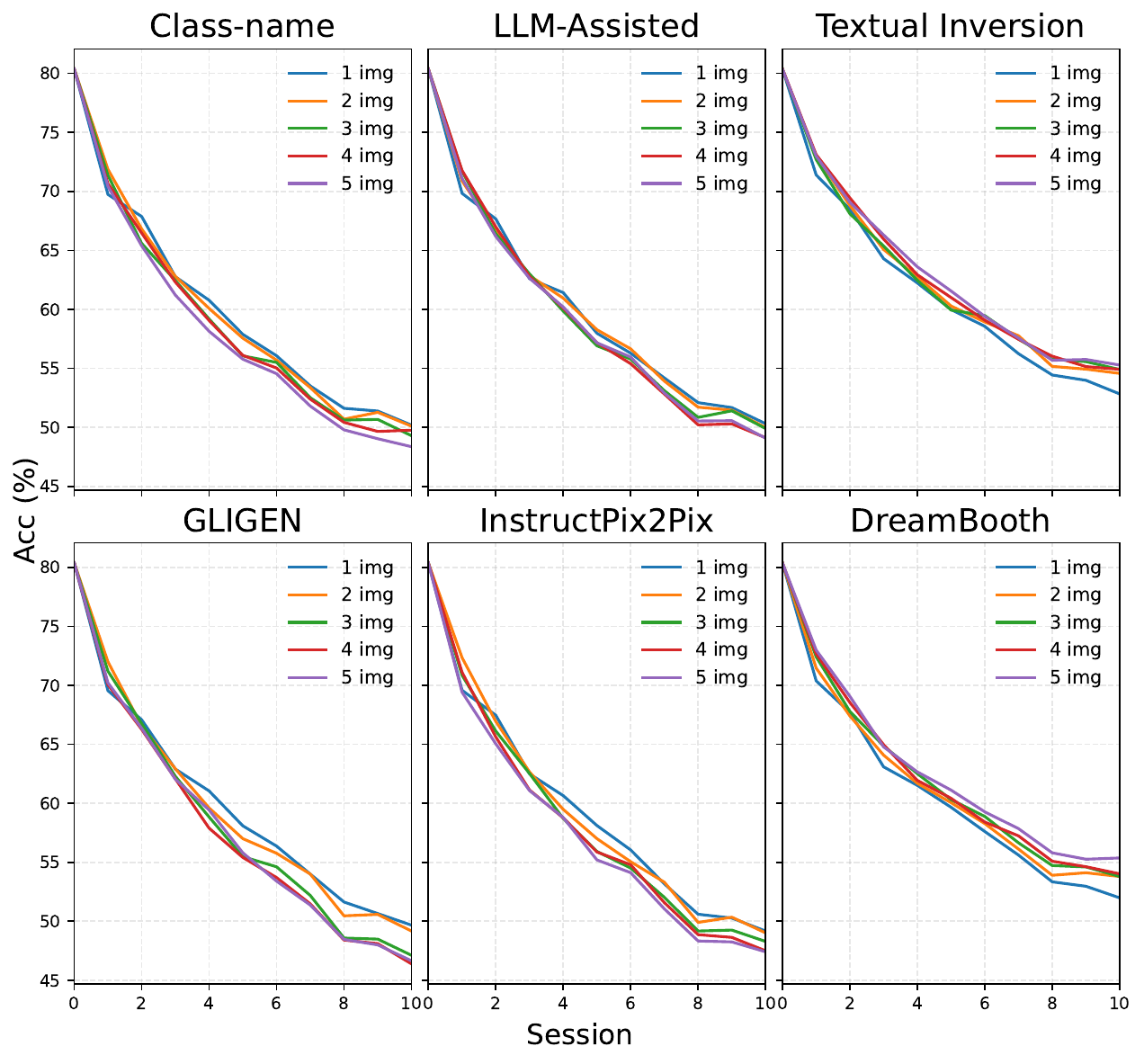}
    \vspace{-1em}
    \caption{\textbf{Effect of synthetic image quantity on CUB-200.} X-axis: session index ($0$–$8$); Y-axis: Top-1 accuracy~(\%). All methods share the same base model.}
    \label{fig:imagecount_cub}
    \vspace{-1em}
\end{figure}

\begin{figure}[t]
    \centering
    \includegraphics[width=\linewidth]{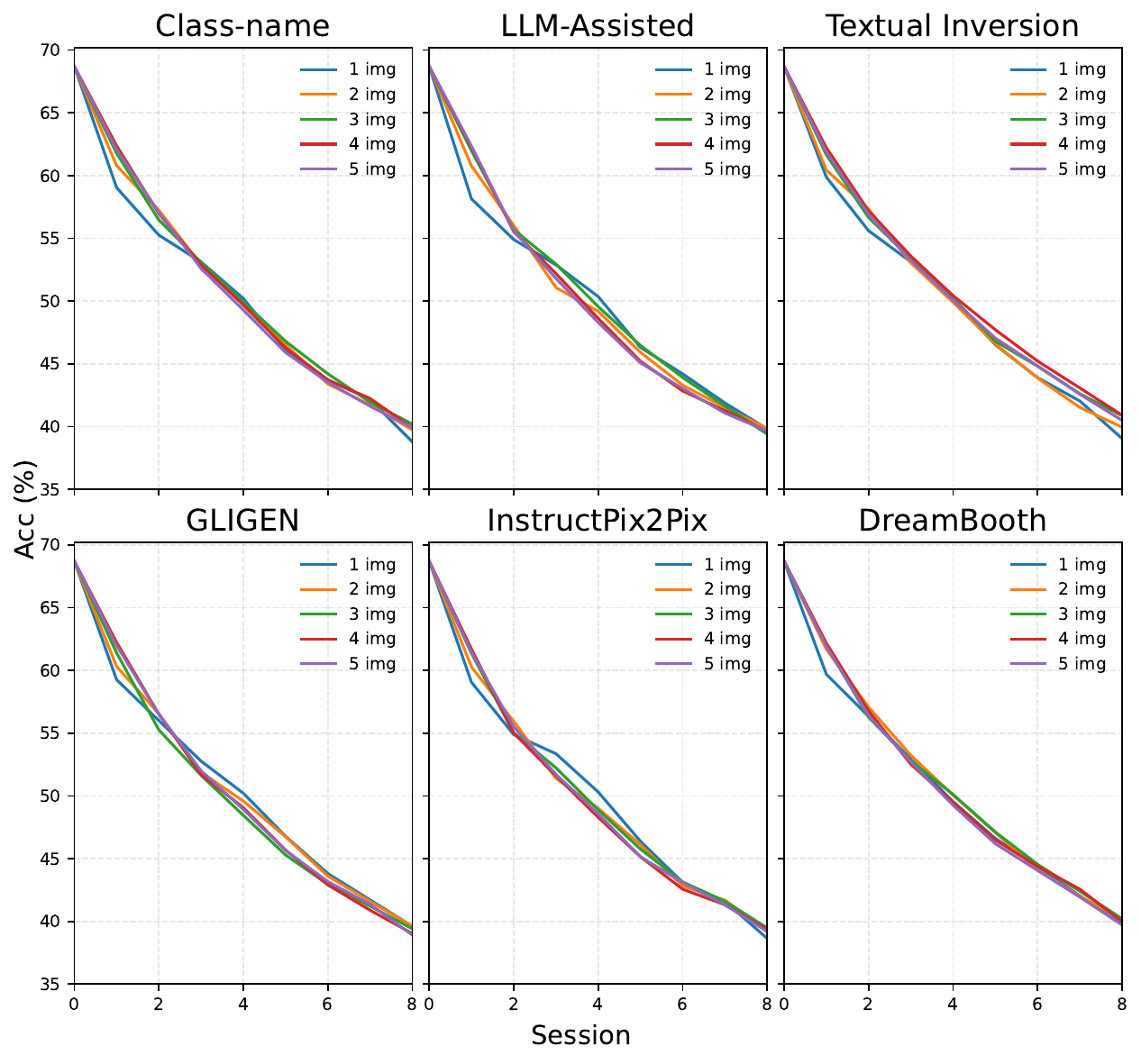}
    \vspace{-1em}
    \caption{\textbf{Effect of synthetic image quantity on \textit{mini}ImageNet.} X-axis: session index ($0$–$8$); Y-axis: Top-1 accuracy~(\%). All methods share the same base model.}
    \label{fig:imagecount_mini}
    \vspace{-1em}
\end{figure}

To clearly understand how the number of synthetic images used as replay $\mathcal{B}_\text{gen}$ affects FSCIL performance, we analyze accuracy trends across incremental sessions for the CUB-200 (\cref{fig:imagecount_cub}) and \emph{mini}ImageNet (\cref{fig:imagecount_mini}) datasets. Specifically, we incrementally increase the number of synthetic replay images per class (from 1 to 5), evaluating multiple generative replay strategies under identical experimental conditions.

For the CUB-200 (\cref{fig:imagecount_cub}), which benefits from richer representations due to ImageNet pretraining, optimization-based methods (\textbf{Textual Inversion}, \textbf{DreamBooth}) consistently outperform others, demonstrating stable or improved accuracy with increased synthetic exemplars. This improvement likely originates from their explicit, class-wise optimization, effectively embedding abundant semantic information within the synthetic images. Conversely, both learning-based methods (\textbf{GLIGEN}, \textbf{InstructPix2Pix}) and prompt-based methods (\textbf{Class-name}, \textbf{LLM-assisted}) exhibit declining performance as the number of synthetic images increases. Such results suggest that merely increasing synthetic data quantity without embedding sufficient class-specific semantic context does not effectively mitigate forgetting, even with high-quality synthetic images.

\begin{figure}[h]
    \centering
    \includegraphics[width=1.00\linewidth]{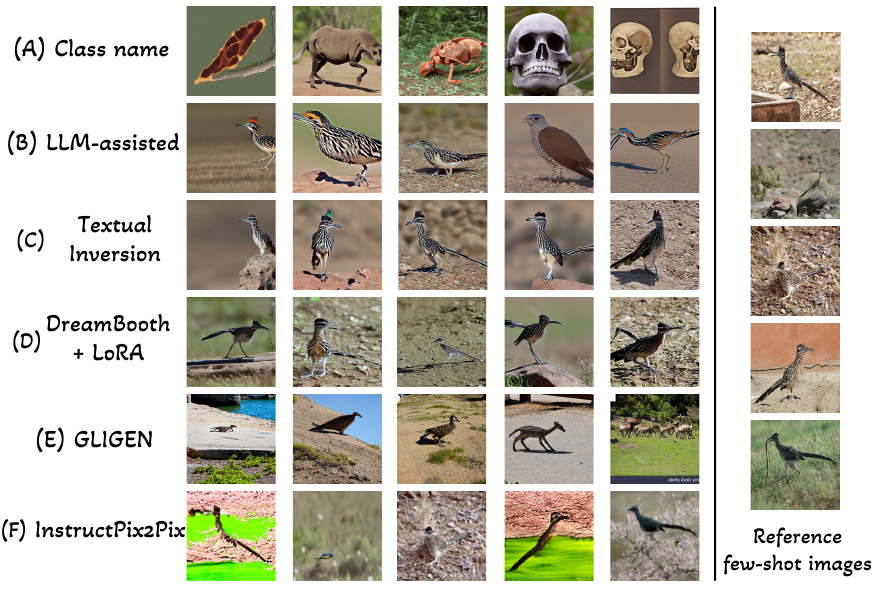}
    \vspace{-1.2em}
    \caption{\textbf{Qualitative results of generation strategies on CUB-200.} We show synthetic images aligning with (A) Class-name, (B) LLM-assisted, (C) Textual Inversion, (D) DreamBooth with LoRA, (E) GLIGEN, (F) InstructPix2Pix, and (rightmost) Reference images used during generation for methods (C), (D), (E), and (F). Class:\small\texttt{Geococcyx} (class 110).}
    \label{fig:qualitative-cub}
    \vspace{-1.5em} 
\end{figure}

\begin{figure}[h]
    \centering
    \includegraphics[width=1.00\linewidth]{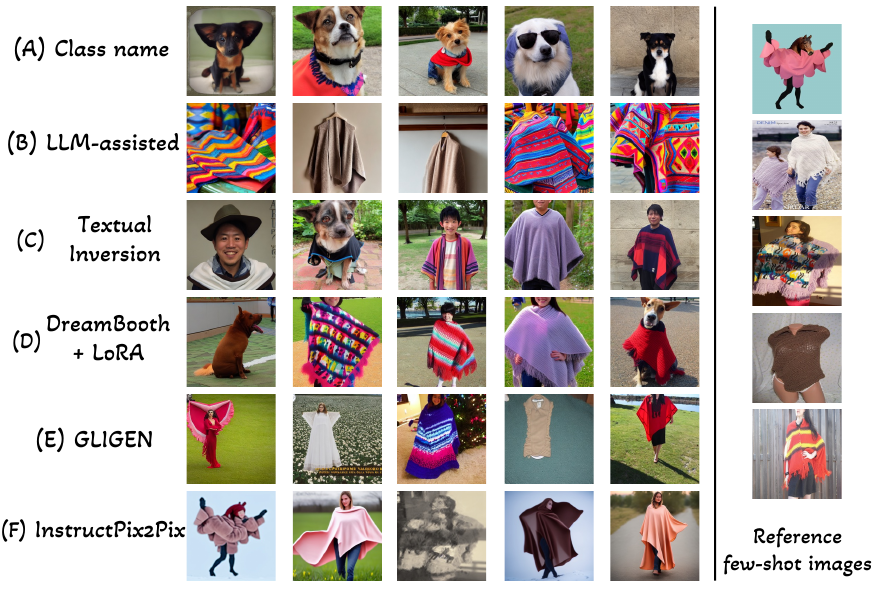}
    \vspace{-1.2em}
    \caption{\textbf{Qualitative results of generation strategies on \textit{mini}ImageNet.} We show synthetic images aligning with (A) Class-name, (B) LLM-assisted, (C) Textual Inversion, (D) DreamBooth with LoRA, (E) GLIGEN, (F) InstructPix2Pix, and (rightmost) Reference images used during generation for methods (C), (D), (E), and (F). Class:\small\texttt{poncho} (class 72).}
    \label{fig:qualitative-mini}
    \vspace{-1.5em} 
\end{figure}
On the other hand, the accuracy variations on \emph{mini}ImageNet (\cref{fig:imagecount_mini}) are less evident. This reduced sensitivity primarily results from limited representational capacity established during the base session and inherent dataset constraints—low image resolution and multiple objects per image—conditions under which SD typically struggles. Nevertheless, \textbf{Textual Inversion} still reveals slight accuracy improvements with increased exemplars, reinforcing the importance of embedding strong class-specific semantic knowledge even in challenging settings. 

These findings highlight that the impact of synthetic replay images in FSCIL strongly depends not just on their quantity but critically on the semantic embedding of each class.

\subsection{Comparative analysis of generative strategies}
\label{subsection:comparison_replay_strategies}
We further compare the effectiveness of different generative replay strategies by fixing the number of synthetic images per class to 5. In \cref{tab:cub_acc} (CUB-200), optimization-based methods (\textbf{Textual Inversion}, \textbf{DreamBooth}) consistently achieve the highest accuracies, with \textbf{Textual Inversion} notably demonstrating the best final-session accuracy (55.28\%) and AA (63.42\%). Conversely, simpler replay methods, such as naïve (\textbf{Class-name}: 48.36\%) and learning-based approaches (\textbf{GLIGEN}: 47.67\%), yield lower final-session accuracy than the baseline without replay (49.64\%). This indicates insufficient alignment with target-domain semantics due to their lack of explicit class-specific embedding mechanisms.

In contrast, \cref{tab:mini_acc} (\emph{mini}ImageNet) reveals a slightly different pattern, with all synthetic replay methods marginally outperforming the baseline without replay (38.40\%). Optimization-based methods remain superior, with \textbf{Textual Inversion} achieving the highest last-session accuracy (40.50\%) and AA (51.75\%). These gains, albeit modest (about 2\%), underscore the importance of explicit, class-specific optimization even under challenging conditions.

Qualitative comparisons (\cref{fig:qualitative-cub,fig:qualitative-mini}) further highlight these differences. \textbf{Textual Inversion} consistently generates images closely resembling reference images, preserving class identity and visual diversity. Conversely, simpler methods like class-name prompting occasionally generate entirely unrelated classes, emphasizing weaker semantic alignment. Although \textbf{DreamBooth+LoRA} generally performs well, it still occasionally misses critical visual nuances, highlighting inherent limitations even in optimized methods.

Overall, our comparative analysis clearly emphasizes the necessity of embedding abundant class-specific semantics within synthetic replay strategies to mitigate catastrophic forgetting effectively, while highlighting existing limitations and avenues for future improvement.

\subsection{Synthetic image timing}
\label{subsection:synthetic_timing}
Enhancing backbone representational capacity during base-session training is critical in FSCIL, as the backbone remains fixed thereafter. Numerous studies~\cite{ahmed2024orco, song2023SAVC, yang2023NC_FSCIL, zhou2022FACT} have aimed to enrich base session representations, typically relying on basic data augmentation (\eg, cutmix~\cite{yun2019cutmix}, mixup~\cite{zhang2017mixup}, \etc). Inspired by recent research~\cite{kim2025diffusionfscil} demonstrating that synthetic images effectively enhance backbone representational capacity, we explicitly investigate the previously unexplored aspect of optimal timing for integrating synthetic images within base session training.

\noindent\textbf{Specific setup for timing analysis.}
As illustrated in~\cref{tab:experiment_settings}, we propose four different schedules for integrating synthetic images during base training: using only real images (\textbf{R (full)}), combining real and synthetic images throughout (\textbf{R+S (full)}), training initially with real images and then fine-tuning with synthetic images (\textbf{R (1st) $\rightarrow$ R+S (2nd)}), and vice versa (\textbf{R+S (1st) $\rightarrow$ R (2nd)}). All scenarios share the same epochs and initial learning rates, with the second stage employing a reduced learning rate (0.1×) for fine-tuning purposes. Synthetic images for base training (30 per class) are generated separately from replay images (5 per class), both produced via the \textbf{Textual Inversion}, with distinct seeds.

\noindent\textbf{Impact of synthetic timing on FSCIL performance.}
Quantitative results are summarized in \cref{tab:ti_timing_compact}. On CUB-200, the two-stage integration strategy (\textbf{R (1st) $\rightarrow$ R+S (2nd)}) achieves the highest Base accuracy (77.93\%), Last accuracy (44.67\%), and AA (57.46\%). However, its relatively high PD (33.27\%) indicates enhanced representational diversity but limited effectiveness at preventing catastrophic forgetting. Conversely, entire integration of synthetic images (\textbf{R+S (full)}) achieves the lowest PD (31.77\%), despite lower starting points (75.80\%). This suggests that frequent exposure to class-aligned synthetic images helps stabilize representations across incremental sessions.

On \textit{mini}ImageNet, the purely real-image scenario (\textbf{R (full)}) consistently yields the best overall results (Base 68.70\%, Last 40.50\%, PD 28.20\%), indicating that synthetic images might introduce detrimental domain shifts, particularly for datasets with smaller resolutions and visually complex scenarios.

\noindent\textbf{Analysis and insights.}
Our findings reveal a clear trade-off between improving representational diversity and effectively mitigating catastrophic forgetting. Thus, carefully considering dataset-specific characteristics to optimally balance synthetic and real images is crucial for enhancing both overall FSCIL performance (AA, Last accuracy) and minimizing performance drop (PD).
\begin{table}[t]
\centering
\caption{FSCIL accuracy (\%) with \textbf{Textual Inversion} under 4 base-session scheduling strategies.
``Base” denotes initial accuracy ($\mathcal{S}_0$), “Last” denotes final accuracy (CUB-200: $\mathcal{S}_{10}$, \textit{mini}: $\mathcal{S}_8$), ``AA” = average accuracy, ``PD” = Base – Last. \textbf{Bold} denotes best.}
\label{tab:ti_timing_compact}
\setlength{\tabcolsep}{3pt}
\renewcommand{\arraystretch}{1.0}

\resizebox{0.98\linewidth}{!}{%
\begin{tabular}{l|cccc|cccc}
\toprule
\multirow{2}{*}[-0.6ex]{\hspace{1em}\makecell[c]{\textbf{Base-session}\\\textbf{data scheduling}}}  %
& \multicolumn{4}{c|}{\textbf{CUB-200}}
& \multicolumn{4}{c}{\textbf{\textit{mini}ImageNet}}\\   %
\cmidrule(lr){2-5}\cmidrule(lr){6-9}
& Base {$\uparrow$} & Last {$\uparrow$} & AA {$\uparrow$} & PD {$\downarrow$}
& Base {$\uparrow$} & Last {$\uparrow$} & AA {$\uparrow$} & PD {$\downarrow$}\\  %
\midrule
\textbf{R}\,+\,\textbf{S} \textit{(full)} & 75.80 & 44.03 & 56.21 & \textbf{31.77} & 68.50 & 39.91 & 51.57 & 28.59\\  %
\textbf{R} \textit{(full)}               & 75.56 & 43.18 & 55.50 & 32.38 & \textbf{68.70} & \textbf{40.50} & \textbf{51.75} & \textbf{28.20} \\
\textbf{R}\,+\,\textbf{S} (1st) $\rightarrow$ \textbf{R} (2nd) & 77.41 & 44.48 & 57.36 & 32.93 & 68.35 & 40.00 & 51.38 & 28.35\\
\textbf{R} (1st) $\rightarrow$ \textbf{R}\,+\,\textbf{S} (2nd) & \textbf{77.93} & \textbf{44.67} & \textbf{57.46} & 33.27 & 68.37 & 39.20 & 50.92 & 29.17\\
\bottomrule
\end{tabular}}
\vspace{-0.6em}
\end{table}

\section{Discussion} 
Our extensive analysis re-examines several previously under-explored aspects of synthetic image replay in FSCIL, including the quantity of synthetic images per class, the choice of generative strategy, and especially the timing of synthetic image integration during base-session training. By systematically analyzing these subtle yet impactful factors, we provide clearer insights into how dataset characteristics influence the effectiveness of synthetic replay methods. Optimization-based methods, particularly Textual Inversion, demonstrate clear advantages in embedding class-specific semantic details, effectively mitigating catastrophic forgetting compared to simpler generative approaches, as quantitatively supported by session-wise accuracy improvements (see Tables~\ref{tab:cub_acc} and~\ref{tab:mini_acc}). Additionally, synthetic integration timing during base training yields distinct, dataset-dependent outcomes, highlighting nuanced trade-offs between representational diversity and forgetting mitigation.

\noindent\textbf{Limitations.} Nonetheless, we observed inherent limitations of SD, particularly its difficulty in fully replicating the complexity and fidelity of real images, thus restricting its practical applicability. Furthermore, to rigorously isolate synthetic image effects, we intentionally maintained fixed training hyperparameters (\eg, learning rate, iteration count), despite recognizing that adapting these parameters could potentially enhance performance. Lastly, unlike typical CL methods~\cite{kim2024sddgr, meng2024diffclass, jodelet2023DiffusionImageReplayCIL, kim2025diffusionfscil} that employ additional domain-alignment techniques (\eg, distillation, regularization, additional losses) when using synthetic replay, our analyses omitted these additional methods, limiting direct comparability with state-of-the-art methods. Through rigorously controlled experiments, our work carefully examines subtle and often overlooked impacts of synthetic images in FSCIL, providing trustworthy insights that enable future research to confidently focus on more significant advancements.

\section{Conclusion}
\label{sectin:conclusion}
In this work, we systematically investigated several subtle yet impactful factors influencing synthetic image effectiveness in FSCIL, including synthetic image quantity, generative strategy, and integration timing during base-session training. Through rigorously controlled experiments on the CUB-200 and \textit{mini}ImageNet datasets, we demonstrated that optimization-based methods, particularly Textual Inversion, significantly outperform simpler generative methods by effectively embedding detailed class-specific semantic information. Additionally, our analysis of synthetic image integration timing uncovered nuanced, dataset-dependent trade-offs between representational diversity and catastrophic forgetting mitigation. Despite the inherent limitations of current generative models, our results provide trustworthy and foundational insights into previously overlooked aspects, paving the way for future research to confidently address more substantial challenges in synthetic-image-driven FSCIL.

\section{Acknowledgment}
This research was supported by Brian Impact Foundation, a non-profit organization dedicated to the advancement of science and technology for all.

{
    \small
    \bibliographystyle{ieeenat_fullname}
    \bibliography{main}
}

\end{document}